\documentclass{article}
\usepackage{ijcai20}

\usepackage{times}

\usepackage{soul}
\usepackage{url}
\usepackage[hidelinks]{hyperref}
\usepackage[utf8]{inputenc}
\usepackage[small]{caption}
\usepackage{graphicx}
\usepackage{amsmath}
\usepackage{booktabs}
\usepackage{graphicx}  
\usepackage{mathrsfs}
\usepackage{amsmath}
\usepackage{amssymb}
\usepackage{algorithm}
\usepackage{algorithmic}

\usepackage{multirow}
\usepackage{booktabs}

\usepackage{color}
\usepackage[T1]{fontenc}
\usepackage{amsmath}
\DeclareMathOperator*{\argmax}{argmax}
\DeclareMathOperator*{\argmin}{argmin}

\title{SparseGAN: Sparse Generative Adversarial Network for Text Generation}

\author{
Liping Yuan$^{1}$ \and
Jiehang Zeng$^{1}$\and
Xiaoqing Zheng$^{1}$ \and
Jun He$^{2}$
\\
\affiliations
$^{1}$Fudan University \and $^{2}$the Administrative Center of Shanghai R\&D Public Service Platforms\\
\emails
\{lpyuan19, jhzeng18, zhengxq \}@fudan.edu.cn,
jhe@sgst.cn,
}

\begin{document}

\maketitle
\begin{abstract}
It is still a challenging task to learn a neural text generation model under the framework of generative adversarial networks (GANs) since the entire training process is not differentiable. 
The existing training strategies either suffer from unreliable gradient estimations or imprecise sentence representations. 
Inspired by the principle of sparse coding, we propose a SparseGAN that generates semantic-interpretable, but sparse sentence representations as inputs to the discriminator.
The key idea is that we treat an embedding matrix as an over-complete dictionary, and use a linear combination of very few selected word embeddings to approximate the output feature representation of the generator at each time step. 
With such semantic-rich representations, we not only reduce unnecessary noises for efficient adversarial training, but also make the entire training process fully differentaiable.
Experiments on multiple text generation datasets yield performance improvements, especially in sequence-level metrics, such as BLEU.
\end{abstract}

\section{Introduction}

Text generation is an important task in natural language processing. Recurrent neural networks (RNNs) have been empirically proven to be quite successful for text generation task due to their capability to capture long-range dependencies. 
By far the most popular strategy to train RNNs is maximum likelihood estimation (MLE), which maximizes the probability of the next word in a sequence given the current (recurrent) state and previous ground truth word (also known as teacher forcing).
At inference time, truth previous words are unknown, and then are replaced by words predicted by the model itself.
The models trained by the teacher forcing strategy usually suffer from the discrepancy between training and inference, called exposure bias \cite{ranzato2015sequence}, which yields errors because the model is only exposed to distribution of training data, instead of its own prediction at inference time.


Most recently, generative adversarial networks (GANs) \cite{goodfellow2014generative} have been used to deal with the exposure bias of RNNs \cite{yu2017seqgan,che2017maximum,lin2017adversarial}. In a typical GAN-based text generation framework, a generator is used to generate sentences given random inputs, and a discriminator is trained to distinguish natural sentences from the generated ones. The generator and discriminator play in a two-player game, and such competition drives the both to improve their desired performance. 

Even though GAN-based approaches have shown to be promising for text generation task \cite{ke2019araml,nie2019ICLR}, it is still challenge to train a GAN-based text generation model due to the discrete nature of text. The output of the generator will be sampled to generate discrete texts, which results in a non-differentiable training process because the gradients can not back-propagate from the discriminator to the generator. Reinforcement learning (RL) technique was introduced to handle the non-differentiable issue \cite{yu2017seqgan,che2017maximum,fedus2018maskgan}, but it still suffers from high-variance gradient estimates, which is hard to alleviate \cite{li2019adversarial}.

An alternative to deal with the non-differentiable issue is to use a continuous function to replace the samplings. After a multinomial distribution over the words from a given vocabulary is estimated by the generator, a differentiable sample, Gumbel Softmax for example \cite{jang2016categorical}, that can be smoothly annealed into a categorical distribution is used to replace the non-differentiable sample from a categorical distribution.
However, as the support set of the multinomial distribution is the whole vocabulary, words with close-to-zero probabilities are all taken into consideration. 
Such approximation become imprecise since these unnecessary words account for a large majority of the vocabulary, which is well-known as the long-tailed distribution. 
Although it can be mitigated via temperature to control the ``steepness'' of the distribution, this problem cannot be completely solved because many unwanted words with nonzero probabilities are still involved, which makes the training inefficient.

To address the above problem, we propose a SparseGAN that generates low-noise, but semantic-interpretable, sparse distributions (i.e. convex combinations of very few word embeddings) to replace the non-differentiable sample. 
With such semantic-rich representations, we not only reduce unnecessary noises for efficient adversarial training, but also make the entire training process fully differentaiable.
Sparse representation has been proven to be powerful for compressing high-dimensional signals \cite{huang2007sparse}. It is used to search for the most compact representation of a signal in terms of the linear combination of several signals in an overcomplete dictionary.

In the SparseGAN, we take the entire word embedding matrix as an overcomplete dictionary, and form the sparse representations as the convex combinations of just a few word embeddings. Those sparse representations are concatenated and fed into a CNN-based discriminator. We also show that such sparse representations can be produced by a matching pursuit algorithm \cite{mallat1993matching}. Generally speaking, no matter what neural network architectures are used in NLP, semantic feature representations at each layer are derived from the input (word) embeddings. Our approach encourage the generator and the discriminator in the GAN-based framework to share the same input feature space spanned by the word embeddings, which can be viewed as a regularization facilitating network training and yielding the better performance.

\section{Related work}
\textbf{GAN-baserd Text Generation} There are mainly two methods to train GAN-based text generation models with the non-differentiable issue caused by the discrete data nature. 
One is to use the RL algorithm, another is to introduce a continuous function to approxijjate the discrete data in latent space.

RL-based GANs usually treat the generator as an agent, where states are the generated words so far and actions are the next words to be generated. Specifically, SeqGAN \cite{yu2017seqgan} models text generation  by sequential decision making process and trains the generator with the policy gradient algorithm. MaliGAN \cite{che2017maximum} trains GAN with maximum likelihood objective to reduce the gradient variance. RankGAN \cite{lin2017adversarial} introduces a margin-based ranking classifier as the discriminator instead of the original binary classifier. LeakGAN \cite{guo2018long} allows the discriminator to leak its own high-level features to the generator to counter the sparse signal from the discriminator. MaskGAN \cite{fedus2018maskgan} introduces an actor-critic conditional GAN that fills in missing text conditioned on the surrounding context to improve sample quality. However, RL-based models usually suffer from large variance of gradient estimation and are difficult to converge.

An alternative method is to approximate the discrete data in the continuous latent space to deal with the non-differentiable problem. WGAN \cite{gulrajani2017improved} feeds the multinomial distribution produced by the generator directly to the discriminator to avoid the sampling operations. GSGAN \cite{jang2016categorical} applies Gumbel-Softmax trick to re-parameterize a discrete distribution, which provides a differentiable way to sample from discrete random variables. RelGAN \cite{nie2019ICLR}, TextGAN \cite{zhang2017adversarial}, GAN-AEL \cite{xu2017neural} use a weighted sum over the embeddings matrix to yield an approximate representation of the generated word sequences, where the weight is the probability of the corresponding word in multinomial distribution. These models confine the inputs of the discriminators to the feature space spanned by the word embeddings. Since the embedding matrix is shared by the generated sentences and real sentences, it will be easier for the discriminator to converge. 
However, these methods suffer from long-tail problem due to the large size of a vocabulary, resulting imprecise approximation of the discrete data. 
Another type of GANs directly work in latent space derived from the generator or the encoder of the auto-encoder. GAN2vec \cite{budhkar2019generative} generates real-valued word2vec-like vectors as opposed to discrete multinomial distribution during training.
ARAE \cite{junbo2017adversarially} combines auto-encoder with GANs for text generation, where the intermediate representations of the auto-encoder are directly used for adversarial training.
Since the latent spaces of generated sentences and real ones are usually different, it can be difficult to minimize the distance between them.

\begin{figure}[t]
 \small
 \centering
 \includegraphics[width=8cm]{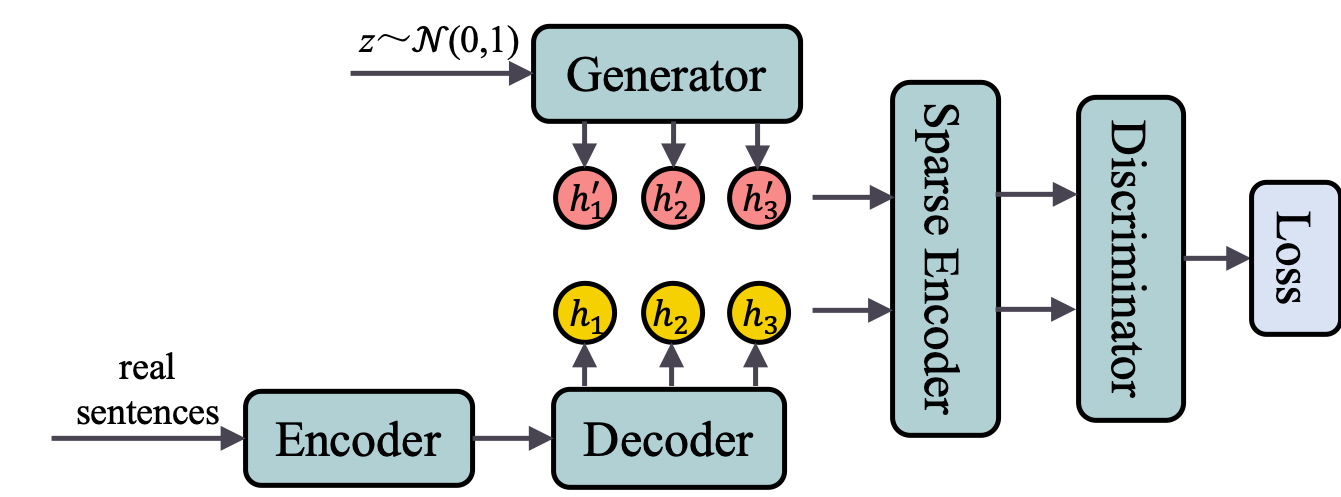}
 \caption{Architecture. The sentence feature representations at each step produced by the generator and the auto-encoder are transformed into their sparse representations by the sparse encoder. Those sparse representations are then summarized and fed into the discriminator to determine whether the sentences are natural or generated ones. The gradients derived from the discriminator's predictions back-propagating to all previous states in an end-to-end manner. By the sparse representations, just a few words are involved in parameter updates that restricts unnecessary noises and facilitates the training.}
 \label{fig:M_archi}
\end{figure}

\textbf{Sparse Representation}
The notion of sparse representation was proposed by Mallat et al \shortcite{mallat1993matching}. The core idea of sparse representation is to approximate a signal in terms of a linear combination of  some selected basis elements from a prespecified dictionary. To extract appropriate basis elements, various optimization algorithms have been applied, such as greedy algorithm and convex relaxation. Some examples of greedy algorithm include Matching Pursuit (MP), Orthogonal Matching Pursuit (OMP) \cite{tropp2007signal}, and Compressive Sampling Matching Pursuit (CoSAMP) \cite{needell2009cosamp}. Convex relaxation is another kind of algorithm to solve the sparse signal representation problem, including Basis Pursuit (BP) \cite{chen2011hyperspectral}, Gradient Projection for Sparse Reconstruction (GPSR) \cite{figueiredo2007gradient}, and Gradient Descent (Grades) \cite{garg2009gradient}. Sparse representation has achieved great success in computer vision, such as face recognition \cite{wright2008robust} and object detection \cite{he2016robust}, but has drawn relatively little attention in NLP. To the best of our knowledge, SparseGAN is among the first ones that incorporate the idea of sparse representation into GAN-based text generation task.

\section{Model}

We here describe the proposed SparseGAN for discrete text generation. As shown in Figure \ref{fig:M_archi}, the SparseGAN consists of four components: a generator $G_{\theta}$ to generate sentences, an auto-encoder to extract the latent representation of real sentences, a sparse encoder for rendering sparse representations, and a discriminator $D_{\phi}$ to distinguish real sentences from the generated ones, where $\theta$ and $\phi$ are model parameters. 

\subsection{LSTM-based Generator}
During adversarial training, the generator takes a random variable $z$ as input, and outputs the latent representation of generated sentence $H_g \in \mathbb{R}^{T \times d}$ using a multi-layer Long Short-Term Memory (LSTM) decoder \cite{schmidhuber1997long}: 
\begin{equation}
    H_g = G_\theta(z)
\end{equation}
where $T$ denotes the sentence length and $d$ the dimensionality of hidden states.

Specifically, the random variable $z$ has a standard normal distribution $z \sim \mathcal{N}(0,1)$ that is taken as the initial value of the LSTM decoder's hidden state. Then, at each time stamp $t$, the LSTM decoder outputs the hidden state $h_t \in \mathbb{R}^{d}$ given previous state $h_{t-1} \in \mathbb{R}^{d}$ and previous word $v_{t-1} \in \mathbb{R}^{d}$ predicted by the model:
\begin{equation}
    h_t = \mathcal{H}(h_{t-1},v_{t-1})
\end{equation}
where $\mathcal{H}(\cdot, \cdot)$ is the standard forward process of a LSTM decoder. 
Once the whole sequence is generated, the sentence representation $H_g$, is derived as the concatenation of all hidden states:
\begin{equation}
    H_g = [h_1, h_2,..., h_T]
\end{equation}
where $[\cdot , \cdot]$ denotes the concatenation operation of multiple vectors. 

\subsection{Denoising Auto-encoder}
The purpose of introducing a pretrained denoising auto-encoder (DAE) \cite{vincent2008extracting} into the GAN-based text generation model is to force the generator to mimic the reconstructed latent representations $H_r \in \mathbb{R}^{T \times d}$ of real sentences instead of the conventional embedding representations \cite{haidar2019latent}. The DAE consists of two parts: a multi-layer bi-LSTM encoder to encode the input real sentence $r$ into intermediate representation, and a multi-layer LSTM decoder to decode the reconstructed hidden state $h^{'}_t \in \mathbb{R}^{d}$ at each time stamp. Similar to the generator, these hidden states $h^{'}_t$ are concatenated jointly to form the latent representation $H_r \in \mathbb{R}^{T \times d}$ of the real sentence $r$.

\subsection{Sparse Encoder}
The role of the sparse encoder is to provide a sparse version of the sentence representation including the generated sentence's representation $H_g$ output by generator and the real sentence's representation $H_r$ output by DAE: 
\begin{equation}
\label{con:sparse_formula}
\begin{aligned}
    S_g = \mathcal{F}_{sparse}(H_g)\\
    S_r = \mathcal{F}_{sparse}(H_r)
\end{aligned}
\end{equation}
where $S_g, S_r \in \mathbb{R}^{T \times d}$, and $\mathcal{F}_{sparse}(\cdot)$ denotes the sparse representation learning algorithm (See Section \ref{sparse}).

\subsection{CNN-based Discriminator}
A commonly used discriminator for text generation is a Convolutional neural network (CNN) classifier which employs a convolutional layer with multiple filters of different sizes to capture relations of various word lengths, followed by a  fully-connected layer. The CNN-based discriminator takes the sparse representation $S \in \mathbb{R}^{k \times b}$ output by the sparse encoder as input , and output a score to determine whether the sentences are natural or generated ones. Formally, the scoring function is defined as follows:
\begin{equation}
    D_{\phi}(S) = W f ( S * \omega ) + b
\end{equation}
where $*$ denotes the convolution operator; $f(\cdot)$ denotes a non-linear function and $W, b , \omega$ are model parameters.

\subsection{Loss Function}
Inspired by Wasserstein GAN (WGAN) \cite{gulrajani2017improved}, the game between the generator $G_\theta$ and the discriminator $D_\phi$ is the minimax objective:

\begin{equation}
\label{con:model_loss}
\begin{aligned}
    L & = \mathbb{E}_{z \sim \mathbb{P}_g}[D_\phi(S_g)] - \mathbb{E}_{r \sim \mathbb{P}_r}[D_\phi(S_r)]\\
        & + \lambda  \mathbb{E}_{\hat{x} \sim \mathbb{P}_{\hat{x}}}[(||\nabla_{S_{\hat{x}}}D_\phi(S_{\hat{x}})||_2-1)^2] 
\end{aligned}
\end{equation}
where $\mathbb{P}_r$ is the data distribution, $\mathbb{P}_g$ is the distribution of the generator's input and $S_g, S_r$ are defined in Equation \ref{con:sparse_formula}. The gradient penalty term \cite{gulrajani2017improved} in the objective function enforces the discriminator to be a 1-Lipschitz function, where $\mathbb{P}_{\hat{x}}$ is the distribution sampling uniformly along straight lines between pairs of points sampled from the $\mathbb{P}_r$ and $\mathbb{P}_g$, while $S_{\hat{x}}$ is the sparse representation of $\hat{x}$ output by the sparse encoder. The importance of this gradient penalty term is controlled by a hyperparameter $\lambda$.

\section{Sparse Representation Learning}\label{sparse}

The sparse encoder aims at finding a sparse equivalence of the sentence representation $H \in \mathbb{R}^{T \times d}$. As described before, $H$ is the concatenation of all hidden states, implying the sparse representation can be computed independently for each state. In this section, we denote $h_t$ as $t$-th state of $H$ for simplicity.

\subsection{Problem Definition}

Sparse representation learning is to search for the most compact representation of a vector via the linear combination of elements in an overcomplete dictionary \cite{mallat1993matching}. Given an overcomplete dictionary $D \in \mathbb{R}^{N \times d}$ with elements in its rows, and a target vector $y\in \mathbb{R}^d $, the problem of sparse representation is to find the sparsest coefficient vector of the linear combination $x^* \in \mathbb{R}^N$ satisfying $y=D^Tx^*$:
\begin{equation}
    x^*=\argmin_{x}{||x||_0}  \quad s.t. \quad y=D^Tx 
\end{equation}
where $||x||_0$ is $\ell_0$-norm of $x$, namely the number of non-zero coordinates of $x$. However, the equality constraint is too strict to be satisfied, and it can be relaxed by minimize the Euclidean distance between $y$ and $D^Tx$. The original problem is then translated into the following problem:
\begin{equation}
\label{con: sparse_definition_math}
    x^*=\argmin_{x}{\lambda ||x||_0 + \frac{1}{2} ||y - D^T x||_2^2} 
\end{equation}
The objective is to reduce the reconstruction error while using the elements as few as possible.
Once the problem solved, $D^Tx^*$ can be used as the final sparse representation of $y$.

Inspired by the sparse representation principle, the sparse encoder takes the vocabulary embedding matrix $E \in \mathbb{R}^{N \times d}$ as the overcomplete dictionary and approximates $h_t$ as the sparse linear combination of word embeddings in $E$, which can be derived as:
\begin{equation}
    c^* = \argmin_{c}{\lambda ||c||_0 + \frac{1}{2} ||h_t - E^T c||_2^2} \label{con:sparse_definition_nlp}
\end{equation}
where $c$ is the coefficient vector of the linear combination. The embedding matrix $E \in \mathbb{R}^{N \times d}$ can be used as the overcomplete dictionary since in text generation tasks, the embedding matrix is always overcomplete with tens of thousands of words, and the condition of $N >> d$ is satisfied in most cases.

\begin{figure}[t]
 \small
 \centering
 \includegraphics[width=7cm]{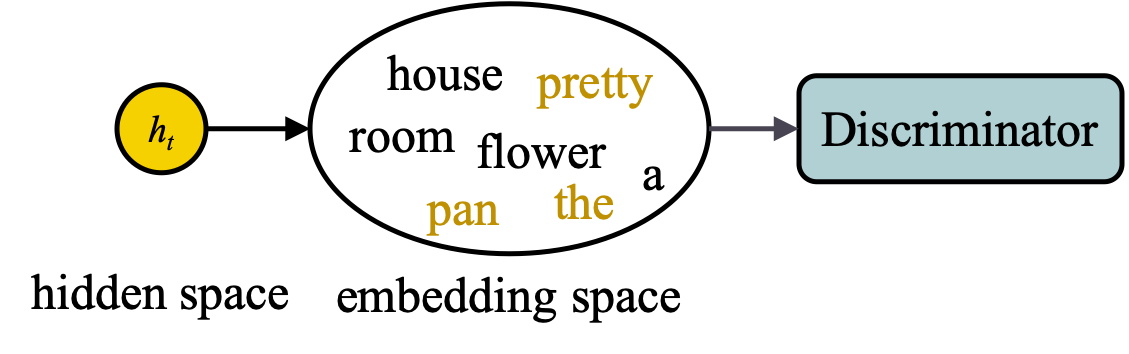}
 \caption{The sparse encoder. It transforms $h_t$ in hidden space into embedding space in an iterative way. The final sparse representations take forms as the linear combinations of just a few word embeddings (indicated by yellow word). }
 \label{fig:M_sparse}
\end{figure}

As shown in Figure \ref{fig:M_sparse}, the constructed sparse representation confines the inputs of the discriminators to the feature space spanned by the word embeddings. Since the generator and DAE share the same embedding matrix, it will be easier for the discriminator to minimize distance between distributions of real sentences and generated ones.
To solve the above optimization problem, we apply the Matching Pursuit (MP) algorithm \cite{mallat1993matching}.

\subsection{Matching Pursuit Algorithm}
The MP algorithm calculates the sparse representation $s_t \in \mathbb{R}^d$ of $h_t$ in an iterative way. As illustrated in Algorithm \ref{alg:sparse_representation}, there is a residual vector $r_t \in \mathbb{R}^d$ to record the remaining portion of $h_t$ that has not been expressed. At a certain iteration $l$, current residue $r_t$ is used to search the nearest word embedding $e_l$ ($l$ represents the $l$-th iteration) from embedding matrix $E$ by comparing the inner product between $r_t$ and all word embeddings in embedding matrix:
\begin{equation}
    e_l = \argmax_{e \in E} \langle r_t, e \rangle 
\end{equation}
where $\langle \cdot, \cdot \rangle$ is the inner product operation of two vectors. The concatenation of $e_l$ and previous selected embeddings forms the basis vector matrix $M \in \mathbb{R}^{k \times d}$, and the linear combination over the row vectors of $M$ is used to approximate $h_t$. The linear combination coefficient vector $c \in \mathbb{R}^k$ is determined by solving the least square problem:
\begin{equation}\label{con:ols}
\begin{aligned}
    c^* &= \argmin_{c}{|| h_t - M^Tc ||_2}\\
    &=  M^+ h_t = (M M^T)^{-1}Mh_t  \\
\end{aligned}
\end{equation}
where $M^{+} \in \mathbb{R}^{k \times d}$ is the pseudo-inverse of $M^T, M^{+} = (M M^T)^{-1}M$. After $c^*$ is calculated, the sparse representation $s_t$ of $h_t$  can be defined as:
\begin{equation}
\label{eqa:sparse_reprensentation}
    s_t = M^Tc^*
\end{equation}
where $s_t$ is the closest to $h_t$ until the current iteration. And $r_t$, the residual vector between $h_t, s_t$ can be defined as:
\begin{equation}
    r_t = h_t - s_t = h_t - M^Tc^{*}
\end{equation}

The process described above will be repeated for $L$ times, where $L$ is a hyperparameter to control the degree of how well $h_t$ is represented approximately. After $L$ iterations, the final sparse representation $s_t \in \mathbb{R}^{d}$ is defined in Equation \ref{eqa:sparse_reprensentation}. For other hidden states $h_1, h_2, ..., h_T \in H$, the same calculation process is performed to obtain their corresponding sparse representations $s_1, s_2, ..., s_T \in \mathbb{R}^b$. These sparse representation are then concatenated together to form the final output $S \in \mathbb{R}^{k \times d}$ of the sparse encoder $S = [s_1, s_2, ... , s_T]$, which is fed into the CNN-based discriminator to determine the score of the input sentence.  

The sparse representation learning algorithm is differentiable. The gradient of $s_t$ can be passed to $c^*$ through Equation \ref{eqa:sparse_reprensentation} and then be passed to $h_t$ through Equation \ref{con:ols}. As a result, SparseGAN is trainable and differentiable via using sstandard back-propagation.



\begin{algorithm}[t]
\caption{Sparse representation learning in SparseGAN}
\label{alg:sparse_representation}
\begin{algorithmic}[1]
\REQUIRE ~~\\
The vector $h_t$; \\
The overcomplete dictionary $E$;\\
Maximum number of iterations $L$;
\ENSURE ~~\\
The sparse representation $s_t$ of $h_t$;
\STATE initial 
$r_t$ = $h_t$;
$M = \varnothing $;
$l=1$;
\REPEAT
\STATE $l=l+1$;
\STATE find $ e_k \in E$ with maximum inner product $\langle e_k, r_t \rangle$;
\STATE $M = [e_1, e_2, ..., e_k]$;
\STATE solve the least square problem,  \\
$c^* = \argmin_{c}{|| h_t - M^Tc ||_2}$;
\STATE compute approximation of $h_t$ by $s_t = M^Tc^*$;
\STATE update residue $r_t = h_t - s_t$;
\UNTIL{$l \ge L$}
\end{algorithmic}
\end{algorithm}

\section{Experiments}

\subsection{Dataset}

We conduct experiments on two different text generation datasets of COCO Image Caption \cite{chen2015microsoft} and EMNLP2017 WMT News \cite{guo2018long} to demonstrate the effectiveness of SparseGAN. 
The COCO Image Caption dataset is preprocessed basically following Zhu et al \shortcite{zhu2018texygen}, which contains $4,682$ distinct words, $10,000$ sentences as train set and other $10,000$ sentences as test set, where all sentences are $37$ or less in length. The EMNLP2017 WMT News dataset contains $5,712$ distinct words with maximum sentence length $51$. The training set and testing set consists of $278,586$ and $10,000$ sentences respectively. 

\begin{table*}[h] \small
	\begin{center}
		\caption{The results of BLEU scores on COCO Image Caption Dataset. BL and SBL denote BLEU and Self-BLEU.}\label{tab:bleu_coco}
		\begin{tabular}{l|c|c|c|c|c|c|c|c}
			\hline
			\hline
			Model & BL2 & BL3 & BL4 & BL5 & SBL2 & SB3 & SBL4 & SBL5  \\
			\hline
			MLE & 0.731	& 0.497	& 0.305	& 0.189	& \textbf{0.916} & \textbf{0.769} & \textbf{0.583} & \textbf{0.408} \\
			SeqGAN \cite{yu2017seqgan} & 0.745	& 0.498	& 0.294	& 0.180	& 0.950	& 0.840	& 0.670	& 0.490 \\
			MaliGAN \cite{che2017maximum} & 0.673	& 0.432	& 0.257	& 0.159	& 0.918	& 0.781	& 0.606	& 0.437 \\
			RankGAN \cite{lin2017adversarial} & 0.743	& 0.467	& 0.264	& 0.156	& 0.960	& 0.883	& 0.763	& 0.619 \\
			LeakGAN \cite{guo2018long} & 0.746	& 0.528	& 0.355	& 0.230	& 0.966	& 0.913	& 0.849	& 0.780 \\
			\hline
			TextGAN \cite{zhang2017adversarial} & 0.593 & 0.463 & 0.277 & 0.207 & 0.942 & 0.932 & 0.805 & 0.746 \\
			\hline
			LATEXTGAN \cite{haidar2019latent} & 0.787 & 0.496 & 0.286 & 0.150 & 0.988 & 0.950 & 0.847 & 0.612 \\
			\hline
			TopKGAN-S & 0.786 & 0.588 & 0.399 & 0.260 & 0.949 & 0.868 & 0.751 & 0.613 \\
			TopKGAN-D & 0.783 & 0.578 & 0.390 & 0.255 & 0.947 & 0.865 & 0.746 & 0.607 \\
			\hline
			SparseGAN & \textbf{0.845} & \textbf{0.666} & \textbf{0.474} & \textbf{0.323} & 0.954 & 0.951 & 0.896 & 0.816 \\
			\hline
			\hline
		\end{tabular}
	\end{center}
\end{table*}

\begin{table*}[h] \small
    \centering
    \caption{The results of BLEU scores on EMNLP2017 WMT News Dataset. BL and SBL denote BLEU and Self-BLEU.}
    \begin{tabular}{l|c|c|c|c|c|c|c|c}
        \hline
        \hline
        Model & BL2 & BL3 & BL4 & BL5 & SBL2 & SB3 & SBL4 & SBL5 \\ \hline
        MLE & 0.749 & 0.453 & 0.223 & 0.113 & \textbf{0.840} & \textbf{0.555} & \textbf{0.301} & \textbf{0.163} \\
        \hline
        SeqGAN \cite{yu2017seqgan} & 0.668 & 0.388 & 0.192 & 0.101 & 0.883 & 0.645 & 0.233 & 0.400 \\
        MaliGAN \cite{che2017maximum} & 0.727 & 0.395 & 0.160 & 0.077 & 0.873 & 0.643 & 0.404 & 0.226 \\
        LeakGAN \cite{lin2017adversarial} & 0.733 & 0.421 & 0.185 & 0.092 & 0.887 & 0.659 & 0.399 & 0.216 \\
        \hline
        TextGAN \cite{zhang2017adversarial} & 0.245 & 0.204 & 0.165 & 0.108 & 0.999 & 0.999 & 0.999 & 0.999 \\ 
        RelGAN \cite{nie2019ICLR} & 0.887 & 0.725 & 0.495 & 0.287 & 0.969 & 0.901 & 0.797 & 0.671 \\
        \hline
        TopKGAN-S & 0.883 & 0.703 & 0.468 & 0.275 & 0.961 & 0.881 & 0.758 & 0.615 \\
		TopKGAN-D & 0.904 & 0.743 & 0.525 & 0.332 & 0.966 & 0.899 & 0.801 & 0.689 \\
        SparseGAN & \textbf{0.921} & \textbf{0.825} & \textbf{0.643} & \textbf{0.437} & 0.992 & 0.982 & 0.961 & 0.930 \\
        \hline
        \hline 
    \end{tabular}
    \label{tab:bleu_emnlp}
\end{table*}

\subsection{Experiment Settings}


The generator is a two layer LSTM with 300 hidden units and the discriminator is a multi-layer 1-D convolution neural network with 300 feature maps and filter size set to 5. The denoising auto-encoder (DAE) is a two layer LSTM with 300 hidden cells for both the encoder and the decoder. For training DAE, we preprocess the input data following Freitag and Roy \shortcite{freitag2018unsupervised}, where $50\%$ of words are randomly removed and all words are shuffled while keeping all word pairs together that occur in original sentence.  
A variational auto-encoer (VAE) \cite{kingma2013auto} is used to initialize the generator, which is trained with KL cost annealing and word dropout during decoding following Bowman et al \shortcite{bowman2015generating}. Inspired by WGAN-GP \cite{gulrajani2017improved}, the hyperparameter $\lambda$ of the gradient penalty term in Equation \ref{con:model_loss} is set to 10, and 5 gradient descent steps on the discriminator is performed for every step on the generator. 
All models are optimized by Adam  with $\beta_1 = 0.9$, $\beta_2 =  0.999$ and $ eps= 10^{-8}$. Learning rate is set to $10^{-3}$ for pretraining and $10^{-4}$ for adversarial training.
The 300-dimensional Glove word embeddings released by Pennington et al \shortcite{pennington2014glove} are used to initialize word embedding matrix.
The batch size is set to $64$, the maximum of sequence length to $40$, the maximum of iterations for adversarial training to $20,000$, and the number of iterations $L$ for sparse representation learning to $10$.

\begin{table}[ht]  \small
    \centering
     \caption{Example sentences generated by SparseGAN trained on COCO Image Caption dataset (shown in the top) and EMNLP2017 WMT News dataset (shown in the bottom) respectively.}
     \label{tab:my_label}
     \begin{tabular}{p{80mm}}
        \hline
        \hline
         a motorcycle is parked on a concrete road . \\
         the picture of a kitchen with stainless and white appliances. \\
         a man riding a motorcycle down a road with a person on the back.  \\
         people are preparing food in a kitchen with a pot. \\
         two teddy bears on a sidewalk next to a baby giraffe. \\
         a table with various cakes and a cup of sausage on it.\\
         an old kitchen with a black and white checkered floor.\\
         a motorcycle is parked on the side of a road with a crowd of people. \\
         a kitchen with hardwood cabinets and white appliances.\\
         a small bathroom with a white toilet and a latticed mirror. \\
         \hline
         i think that's the most important thing that's happening, there is a lot of ideas in the white house of the next time. \\
         the queen's story is aimed on making a positive increase in the uk's population in scotland. \\
         the government's executive ministry said: `` it was just a very positive problem in my relationship and i am pleased to be able to make sure it would be. \\
         `` i think it's going to be investigated, but it doesn't matter , if she can have a child , '' he says. \\
         the queen government is asking to comment on a review of the brexit referendum, and asked whether this was not a big question. \\
         the government also said that's president obama had to do that negotiations and we did not consider the possibility of parliament to be successful, it's not a good team. \\
         `` the first message, to say that trump will be a bitter path to the white house, '' kaine said. \\
         `` it's hard to get a good team, and we don't want to get the best players in the country, '' he said. \\
         it's important that i'm working at the best time in the world , there's diversity of people who are talented guys, '' he said. \\
         there are a lot of people who are going to go on the work , especially on the day, '' pence said. \\
         \hline
         \hline
    \end{tabular}
    \label{tab:samples}
\end{table}

\subsection{Evaluation Metrics}

We use two metrics below to evaluate our models, comparing different models.

\textbf{BLEU} \cite{papineni2002bleu} This metric is used to measure quality of generated sentences. To calculate BLEU-N scores, we generate $10,000$ sentences as candidate texts and use the entire test set as reference texts. The higher the BLEU score is, the higher quality the generated sentences is.

\textbf{Self-BLEU} \cite{zhu2018texygen} This metric is used to measure diversity of generated sentences. Using one generated sentence as candidate text and others as reference texts, the BLEU is calculated for every generated sentence, and the average BLEU score of $10,000$ generated sentences is defined as the self-BLEU. The higher the self-BLEU score is, the less diversity the generated sentences is.

\subsection{Compared Models}
We compared the SparseGAN with several recent representative models. For RL-based text generative models, we choose to compare SeqGAN \cite{yu2017seqgan}, MaliGAN \cite{che2017maximum}, RankGAN \cite{lin2017adversarial} and LeakGAN \cite{guo2018long}. We also compared TextGAN \cite{zhang2017adversarial} and LATEXTGAN \cite{haidar2019latent}. TextGAN adopts the weighted sum over the embeddings matrix as the continuous approximation, while LATEXTGAN uses the multinomial distribution by the generator for adversarial training. 

Top-$k$ method, which has not been applied to GAN-based text generation model, approximates the sampling action via the linear combination of word embeddings with $k$ highest probabilities and use the re-normalized probabilities as the linear combination coefficients. Here we denote this model as TopKGAN-S (static TopKGAN). TopKGAN-D (dynamic TopKGAN), a variant of TopKGAN-S, chooses words dynamically via comparing the logit of each word with a threshold $\delta$. The logit of each word is defined as the inner product between hidden state $h_t$ and word embedding $e_k$. For TopKGAN-S, the number of words to be chosen $K$ is set to 10, while for TopKGAN-D, the threshold $\delta$ is set to 0 here. Two variants of TopKGAN are implemented with the same setting as SparseGAN.

\subsection{Results}
The BLEU scores and Self-BLEU scores on COCO Image Caption dataset and EMNLP2017 WMT News dataset are shown in Table \ref{tab:bleu_coco} and Table \ref{tab:bleu_emnlp}, correspondingly. The proposed SparseGAN achieves the highest BLEU scores on both datasets, which means the generated sentences by SparseGAN is more natural and fluent. 
MLE-based model has the lowest Self-BLEU scores.
TopKGAN-S and TopKGAN-D have similar performance on both COCO Image Caption dataset and EMNLP2017 WMT News dataset. These two models behave better than several competitive models in terms of both BLEU sores and Self-BLEU scores, such as RankGAN, LeakGAN, TextGAN and LATEXTGAN on COCO Image Caption dataset.

High BLEU scores of SparseGAN may benefits from the way to treat the embedding matrix. Since SparseGAN chooses word embedding via the residual vector $r_t$, which is initialized as $h_t$, the word with the highest probability will be chosen at the first iteration. This word is usually a common word in vocabulary. After several iterations, when $h_t$ has been well approximated by the sparse representation, uncommon words tend to be chosen. Both common and uncommon words are adjusted in SparseGAN, thus the embedding matrix obtains sufficient training.
However, RL-based models only choose one word to adjust at each time stamp; TopKGAN only choose words with high probabilities, which is usually the common words, to adjust; continuous approximation methods choose all words to adjust but contain much noise in their approximation, resulting in imprecise gradient values.


Low Self-BELU scores of MLE-based model reflects that the generated sentences via MLE-based training are more diverse than all GAN-based models. It implies that GAN-based models tend to suffer from mode collapse and generate safe but similar sentences. However, the generated sentences of MLE-based model are less natural than GAN-based models, especially on EMNLP2017 WMT dataset which has longer sentences than the other dataset.


Table \ref{tab:samples} shows the sentences generated by SpargeGAN that is trained with COCO Image Caption and EMNLP2017 WMT News datasets respectively. Those examples illustrate that SparseGAN is capable of generating meaningful and natural sentences with a coherent structure.

\section{Conclusion}
Generative adversarial networks have been used for text generation in order to alleviate the discrepancy between training and inference (exposure bias). However, simply applying GANs to the generation task will lead to  a non-differentiable training process that hinders the gradients back-propagating to the generator from the discriminator.
We proposed a fully differentiable training solution that is achieved by feeding the discriminator with semantic-interpretable, but anti-noise sparse sentence representations.
The proposed solution encourages the generator and the discriminator to share the same input semantic feature space formed by the word embeddings --- a regularization method that facilitates network training and improves the performance.
Experiments on multiple text generation datasets showed that the proposed model and training algorithm achieved the best or comparable performance, especially in terms of the BLEU scores and self-BLEU scores, reflecting the enhanced ability in recovering the probability of the whole sequence and improving the diversity in the generated sentences.

\section*{Acknowledgements}
This work was supported by Shanghai Municipal Science and Technology Project (No. 21511102800).

\small {
\bibliography{ijcai20}
\bibliographystyle{named}
}
\end{document}